\documentclass[11pt, twocolumn]{article}

\usepackage{geometry}
\usepackage[skip=0.5\baselineskip]{caption}

\usepackage{wrapfig}
\usepackage{array}
\usepackage{hyperref}
\usepackage{graphicx}
\usepackage[ruled,vlined,linesnumbered,algo2e]{algorithm2e}
\usepackage{times,amsmath,epsfig}
\usepackage{booktabs}
\usepackage{subcaption}
\usepackage{footmisc}
\usepackage{multirow}
\DeclareMathOperator*{\argmax}{argmax}

\usepackage{balance}
\makeatletter
\newcommand\footnoteref[1]{\protected@xdef\@thefnmark{\ref{#1}}\@footnotemark}

\makeatother

\begin{document}

\title{Ultra Efficient Transfer Learning with Meta Update for Cross Subject EEG Classification}

\author{Tiehang Duan, Mihir Chauhan, Mohammad Abuzar Shaikh, Jun Chu and Sargur Srihari
\thanks{Tiehang Duan, Mihir Chauhan, Mohammad Abuzar Shaikh, Jun Chu  and Sargur Srihari are with the Department of Computer Science and Engineering, University at Buffalo, NY 14260, USA (e-mail: \{tiehangd, mihirhem, mshaikh2, jchu6, srihari\}@buffalo.edu).}
}
\date{}

\maketitle

  \thispagestyle{empty}

\begin{abstract}
     The pattern of Electroencephalogram(EEG) signal differs significantly across different subjects, and poses challenge for EEG classifiers in terms of 1) effectively adapting a learned classifier onto a new subject, 2) retaining knowledge of known subjects after the adaptation. We propose an efficient transfer learning method, named \textbf{M}eta \textbf{UP}date \textbf{S}trategy (MUPS-EEG), for continuous EEG classification across different subjects. The model learns effective representations with meta update which accelerates adaptation on new subject and mitigate forgetting of knowledge on previous subjects at the same time. The proposed mechanism originates from meta learning and works to 1) find feature representation that is broadly suitable for different subjects, 2) maximizes sensitivity of loss function for fast adaptation on new subject. The method can be applied to all deep learning oriented models. Extensive experiments on two public datasets demonstrate the effectiveness of the proposed model, outperforming current state of the arts by a large margin in terms of both adapting on new subject and retain knowledge of learned subjects. 
     
     Our code is publicly available at \url{https://github.com/tiehangd/MUPS-EEG}.

\end{abstract}

\section{Introduction}

Electroencephalogram(EEG) signal is widely used to analyze the activities of human brain.
The signal is recorded by placing electrodes on different regions of human scalp when the subject performs executive/imaginary tasks or perceives stimulus from outside\cite{Buzski2012}. 
EEG signal has proved to be effective for restoring motion capabilities of disabled people \cite{Tariq2018}, human intention interpretation \cite{Padfield2019}, and enhanced experience in gaming control \cite{Liao2012}.  

EEG signal exhibit significant pattern variability across subjects, resulting in two major challenges for EEG classifiers: 1) achieve good performance on new users previously unseen, 2) retain knowledge of previous learnt subjects after the adaptation. We propose to simultaneously tackle the challenge with \textbf{M}eta \textbf{UP}date \textbf{S}trategy (MUPS-EEG) involving two steps: (1) extracting versatile features which are effective across different subjects with meta learned representations, and (2) perform meta update for fast adaptation on new subject.
%, denoted as the calibration process\footnote{Calibration-free approaches extract subject independent features without calibration on target. They have the privilege that no additional effort is required from the target subject, but performance improvement is still needed for challenging tasks, as 70\% classification accuracy is generally deemed an acceptable threshold for BCI systems \cite{Fahimi2019}.}.
The meta update mechanism significantly reduced the amount of labeled target data needed to adapt on target subject, and the meta learned representations help preserve learned knowledge on previous subjects. This facilitates the utility of BCI systems in real world scenarios with constant shift between different subjects.

For the extraction of versatile and subject invariant features, previous works adopt either signal processing techniques or deep learning models. For example, \cite{kai2008} utilized filter bank (FB) and common spatial pattern (CSP) for effective feature extraction which are then sent to a fisher linear discriminator (FLD). 
\cite{Jatupaiboon2013} extracted features from power spectral density (PSD) of EEG signals and used support vector machines (SVM) as the classifier.
Models based on deep learning emerged as a promising approach as they alleviate the need for manual feature engineering and achieved state of the art performance. 
EEGNet \cite{Lawhern_2018} is a compact convolutional neural network (CNN) that can be applied to different BCI paradigms.  \cite{ZhangD2018} introduced a cascade and parallel structure on CNN for improved performance. CRAM \cite{zhangd2019} is proposed recently which adopts LSTM with attention mechanism to help the model focusing on most discriminative temporal features, and achieved promising result.

Transfer learning techniques are utilized in EEG classifiers to transform models onto target subject for improved performance. Previous works involves both classic transfer learning \cite{Wu2016} \cite{Driver2017} and domain adaptation\cite{ZhengW2016}\cite{LanZ2019} to transfer knowledge across subjects.
\cite{Fahimi2019} proposed an inter-subject transfer learning framework built on top of CNN model. 
\cite{ZhengW2016} and \cite{LanZ2019} explored performance of multiple domain adaptation methods including transfer component analysis (TCA-EEG), maximum independence domain adaptation (MIDA-EEG) and information theoretical learning (ITL) for emotion recognition. 
 Deep-Transfer \cite{Tan2018} is a transfer learning framework built on deep CNN-LSTM network to transfer knowledge across subjects. RA-MDRM \cite{Zanini2018} utilized covariance matrix from different subjects and forms a calibration less system suitable for low resource scenarios. 

In this work, we propose a simple and computationally efficient meta updating strategy to tackle cross subject EEG classification, which is applicable to all deep learning oriented classifiers. It allows the EEG classifier to adapt onto a new subject utilizing only a small amount of target data. Furthermore, the model mitigates forgetting that often occurs when transferring a deep learning model to a new context. This \textbf{M}eta \textbf{UP}date \textbf{S}trategy (MUPS-EEG) originates from meta learning \cite{Finn2017}\cite{metatransfer2018}. It involves a meta representation learning phase followed by meta adaptation onto target subject. 
The meta representation learning is performed on the known source subjects which extracts versatile features that are effective across different subjects, and meta adaptation   fits the model onto a new subject through a small number of gradient steps without losing knowledge on known subjects.
A desirable property of the model is that it doesn't overfit even if target data is very limited, allowing it to properly function in low target-resource scenarios.

\section{Methodology} 

MUPS-EEG allows efficient adaptation onto a new subject and simultaneously retain knowledge on known subjects. Its meta learnt representations are broadly effective across different subjects, and meta adaptation fits the model to a new subject with efficient target data usage.
The difference between MUPS-EEG and classic transfer learning lies in both the optimization process and the training mechanism.

For traditional optimization, weights are sequentially updated after each time step, seeking sensible parameters with

\begin{equation} \label{eq1}
\hat{\Theta}=\argmax_\Theta \log p(\Theta|\mathcal{D}_s, \mathcal{D}_t)
\end{equation}

where $\Theta$ is the collection of model parameters, $\mathcal{D}_s$ is training data from source subjects, and $\mathcal{D}_t$ is the small amount of data from target subject. 

MUPS-EEG decomposes the problem into two steps by setting up meta parameters $\Phi$. Given

\begin{equation} \label{eq2}
\log p(\Theta|\mathcal{D}_s, \mathcal{D}_t) =\log \int_{\Phi} p(\Theta|\mathcal{D}_t, \Phi)p(\Phi|\mathcal{D}_s)d\Phi
\end{equation}

Maximizing the log likelihood is approximated to first finding meta parameters that maximizes $\log p(\Phi|\mathcal{D}_s)$

\begin{equation} \label{eq3}
\hat{\Phi} = \argmax_\Phi \log p(\Phi|\mathcal{D}_s)
\end{equation}

Then approximates eq. \ref{eq1} to be

\begin{equation} \label{eq4}
\argmax_\Theta \log p(\Theta|\mathcal{D}_s, \mathcal{D}_t) \approx \argmax_\Theta \log p(\Theta|\mathcal{D}_t, \hat{\Phi})
\end{equation}

 The mechanism can thus be interpreted as helping the model learn a prior of transferable knowledge on the subjects. This prior is later used to infer the posterior parameters in the network after the model sees a small amount of data from the new subject. The prior learned during meta training act as an inductive bias for minimizing the generalization error during evaluation, which allows the EEG classifier to properly functions on the new subject.

During meta training, MUPS-EEG involves interaction between a base learner and a meta learner, each formed with a representation learning network and a prediction learning network.
Representation learning network extracts effective features from raw EEG signal which is then feed to prediction learning network for classification. Both representation learning network and prediction learning network can be arbitrary deep learning models.

The workflow of MUPS-EEG is as follows:

An ensemble of $M$ meta tasks $\mathcal{E}_{meta}=\{\mathcal{T}_1, \mathcal{T}_2,...,\mathcal{T}_M\}$ is created from source dataset $\mathcal{D}_s=\{(x_1,y_1),...,(x_N, y_N)\}$ with a total of $L$ known subjects. Each meta task $\mathcal{T}_{i}=\{(x_{1}^{i}, y_{1}^{i}), ..., (x_{m}^{i}, y_{m}^{i})\}$ contains $m$ data points from $l$ subjects, where $m\ll N$ and $l<L$.

\begin{algorithm2e}[!ht]
\caption{MUPS-EEG}
\label{meta}
\SetAlgoLined
\SetNoFillComment
\SetKwInOut{Input}{Input}
\SetKwInOut{Output}{Output}
\Input{data from source subjects $\mathcal{D}_{s}$, data from target subject $\mathcal{D}_{t}$, base learning rate $\alpha$, meta learning rate $\beta$} 
\Output{optimal meta learned model}

\For{samples in $\mathcal{D}_{s}$}{
       pretrain $\phi$ based on $\mathcal{L}_{\mathcal{D}_{s}}(\phi)$
	}

\While{not done}{
sample a batch of tasks $\{\mathcal{T}_{1\sim K}\}\in \mathcal{E}_{meta}$

\For{meta episode k from 1 to K}{
        Split $\mathcal{T}_{k}$ into $\mathcal{T}_{b}$ and $\mathcal{T}_{m}$
                
        \For{number of base updates}{optimize $\theta$ with $\mathcal{T}_{b}$ by Eq. \ref{eq5}} 
        
        optimize $\{\theta^{*}, \phi^{*}\}$ with $\mathcal{T}_{m}$ by Eq. \ref{eq6}. 
        
        $\{\theta, \phi\} \gets \{\theta^{*}, \phi^{*}\}$
	}

}
\end{algorithm2e}

%%%%%%%%%%%%%%%%%%

Each cycle of meta update is called an episode, including two phases: base learning and meta learning. In each episode, a meta task $\mathcal{T}_{i}$ is sampled from the task pool $\mathcal{E}_{meta}$, with $p$ data points for base learning $\mathcal{T}_{b}$, $q$ data points for meta learning $\mathcal{T}_{m}$ (omitted indexing on $i$ here for conciseness), and $p+q=m$.

MUPS-EEG adopts a two stage optimization approach with two sets of optimizers, one for optimizing base learner and the other for optimizing meta learner. Base learner includes representation learning net parameterized with $\phi$ and prediction learning net parameterized with $\theta$. Meta learner keeps another set of parameters $\{\phi^{*}, \theta^{*}\}$. During initialization, $\{\phi, \phi^{*}\}$ is pretrained to have a warm start, and $\{\theta, \theta^{*}\}$ is randomly initiated. In later episodes, both base learner and meta learner inherit parameter values from meta learner of previous episode.

In base learner, gradient is evaluated with the loss function $\mathcal{L}_{\mathcal{T}_{b}}(\theta, \phi)$ being cross entropy for the classification task. When updating base learner, we only update parameters in prediction learning net, which is

\begin{equation} \label{eq5}
\theta \leftarrow \text{Adam}\Big( \theta, \nabla_{\theta}\mathcal{L}_{\mathcal{T}_{b}}(\theta, \phi), \alpha\Big)
\end{equation}

where $\alpha$ is the learning rate for base optimizer. Here Adam can be replaced by any optimizer functioning on first order gradient. After base learning loop ends, meta task $\mathcal{T}_m$ is applied to get meta gradient $\nabla_{\{\theta, \phi\}}\mathcal{L}_{\mathcal{T}_{m}}(\theta, \phi)$, and parameters of both representation learning net and prediction learning net get updated accordingly

\begin{equation} \label{eq6}
\{\theta^{*}, \phi^{*}\} \leftarrow \text{Adam}\Big( \{\theta^{*}, \phi^{*}\}, \nabla_{\{\theta, \phi\}}\mathcal{L}_{\mathcal{T}_{m}}(\theta, \phi), \beta \Big)
\end{equation}
 
where $\beta$ is the learning rate for meta optimizer. Note this meta optimization is performed over the meta learner, whereas the objective gradient is computed from updated base learner for its gradient descent direction is broadly effective on different subjects.
Meta learner is kept between different episodes and then adapt to target subject during evaluation. The algorithm is outlined in Algorithm \ref{meta}.

\section{Experiments}

We compare our method against current state of the arts on two public datasets, with detailed experiment setting described as below.

\textbf{Dataset:} The proposed model is evaluated on two public datasets, namely BCI competition IV dataset 2a (abbreviated as BCI IV-2a below) \cite{Tangermann2012} \footnote{\url{http://bnci-horizon-2020.eu/database/data-sets}} and DEAP dataset \cite{DEAP} \footnote{\url{https://www.eecs.qmul.ac.uk/mmv/datasets/deap/download.html}}. BCI IV-2a involves 9 subjects doing 4 class motor imaginary tasks. Each subject is tested in two sessions and each session consists 288 trials. Signals are recorded with 22 electrodes at 250Hz sampling rate. DEAP dataset is for emotion recognition, with a total of 32 subjects. 40 trials are recorded for each subject as they watched music videos with different types of arousals. The signal comprises 32 channels at a sampling rate of 512Hz.

% Table generated by Excel2LaTeX from sheet 'MUPS'
\begin{table*}[htbp]
  \centering
  \caption{Comparison of Accuracy and ROC-AUC on target subject for BCI-IV 2a and DEAP dataset. BCI-IV 2a has a total of nine subjects, the models are trained on eight subjects and tested on the subject left out. Similarly, for DEAP one subject is left out for testing and models are trained on the other 31 subjects. Reported result is averaged across all the subjects. The first three models are subject independent and don't use any target subject data. For the other transfer learning approaches we used the same amount of target data (1 minute of EEG recording for BCI-IV 2a and 5 minutes recording for DEAP dataset) for a fair comparison. MUPS-EEG outperforms comparison methods on both datasets with its efficient meta adaptation mechanism.}
    \begin{tabular}{l|ll|ll}
    \toprule
    \multicolumn{1}{c|}{\multirow{2}[4]{*}{Method}} & \multicolumn{2}{c|}{BCI-IV} & \multicolumn{2}{c}{DEAP} \\
\cmidrule{2-5}          & \multicolumn{1}{c}{Accuracy} & \multicolumn{1}{c|}{ROC-AUC} & Accuracy & ROC-AUC \\
    \midrule
    EEGNet \cite{Lawhern_2018} & $0.557\pm 0.063$ & $0.704\pm 0.033$ & $0.459\pm 0.073$ & $0.627\pm 0.044$ \\
    CTCNN \cite{Schirrmeister2017} & $0.523\pm 0.105$ & $0.721\pm 0.061$ & $0.396\pm 0.095$ & $0.603\pm 0.048$ \\
    CRAM \cite{zhangd2019} & $0.632\pm 0.080$ & $0.769\pm 0.043$ & $0.565\pm 0.117$ & $0.731\pm 0.078$ \\
    \midrule
    MIDA-EEG \cite{ZhengW2016} & $0.650\pm 0.056$ & $0.793\pm 0.036$ & $0.536\pm 0.108$ & $0.671\pm 0.07$ \\
    TCA-EEG \cite{LanZ2019} & $0.674\pm 0.073$ & $0.817\pm 0.053$ & $0.552\pm 0.114$ & $0.695\pm 0.067$ \\
    Deep-Transfer \cite{Tan2018} & $0.712\pm 0.065$ & $0.841\pm 0.041$ & $0.638\pm 0.131$ & $0.767\pm 0.064$ \\
    RA-MDRM \cite{Zanini2018} & $0.741\pm 0.059$ & $0.846\pm 0.032$ & $0.614\pm 0.096$ & $0.758\pm 0.059$ \\
    \midrule
    MUPS-EEG  & $\boldsymbol{0.763\pm 0.055}$ & $\boldsymbol{0.859\pm 0.038}$ & $\boldsymbol{0.672\pm 0.063}$ & $\boldsymbol{0.782\pm 0.037}$ \\
    \bottomrule
    \end{tabular}%
  \label{t1}%
\end{table*}%

\textbf{Implementation Details:} The model is implemented with Pytorch. 
We used a three layer convolutional neural network (CNN) similar to EEGNet\cite{Lawhern_2018} as representation learning network, which is compact and versatile across different BCI paradigms. Prediction learning network includes two fully connected layers.
Representation learning network is pretrained on SGD optimizer with learning rate set to 0.01. 
Adam optimizer is adopted during meta training for adaptation of base learner and meta learner, with learning rate set to 0.001. The learning rate is discounted by 0.2 every 5 steps. We run 10 epochs for representation learning pretraining, and 20 epochs for meta training. Each meta episode involves ten iterations of base learner update and one meta update. During the meta episode, one data batch containing 12 sampled meta tasks are feed into the model, and each task is made up with 20 data segments. 10 data segments are used for base update and the other 10 segments for meta update.

We evaluate model on: 1) the performance with the new target subject, 2) knowledge retained from previous learnt subjects. The performance on the new subject is measured with both accuracy and AUC-ROC. And the knowledge retaining ability is measured with the averaged accuracy (Avg. Acc) and averaged ROC-AUC (Avg. RA) across previous subjects after adaptation finishes.

\textbf{Result Analysis:} Model performance on target subject for BCI IV-2a dataset and DEAP dataset are presented in table \ref{t1}. We did a comprehensive comparison to models that perform well on cross subject classification tasks with code publicly available. The first three comparison models (EEGNet, CTCNN, CRAM) don't involve the transfer process and no target data is used\footnote{These three models adopt a more challenging problem setting which justifies their relatively lower performance. \label{fn2}}.  For the other transfer learning approaches, we used the same amount of target subject data (1 minute of EEG recording for BCI-IV 2a and 5 minutes EEG recording for DEAP dataset) to have a fair comparison. For BCI-IV 2a dataset, MUPS-EEG has an improvement of at least 2.2\% on accuracy and 1.3\% on AUC-ROC compared with other models. The classification accuracy varies across individual subjects. MUPS-EEG classified 7 out of 9 subjects to above 70\% accuracy, which is generally deemed an acceptable threshold for application of BCI systems\cite{Fahimi2019}. For DEAP dataset, MUPS-EEG outperforms other approaches by at least 3.4\% in accuracy and 1.5\% in AUC-ROC. This performance improvement comes from MUPS-EEG's ability to rapidly adapt onto the target domain with a small amount of target data.

% Table generated by Excel2LaTeX from sheet 'Sheet2'
\begin{table*}[htbp]
  \centering
  \caption{Comparison of averaged accuracy (Avg. Acc) and averaged ROC-AUC (Avg. RA) on learnt source subjects for BCI IV-2a dataset and DEAP dataset. The training setting is the same as described in Table \ref{t1}, Avg. Acc and Avg. RA are evaluated on the source subjects after adaptation finishes. EEGNet, CTCNN and CRAM are subject independent approaches and not included here, as their performance are the same as reported in Table \ref{t1}. MUPS-EEG performs consistently better than comparison baselines in retaining knowledge of learned subjects. }
    \begin{tabular}{l|ll|ll}
    \toprule
    \multicolumn{1}{c|}{\multirow{2}[4]{*}{Method}} & \multicolumn{2}{c|}{BCI-IV} & \multicolumn{2}{c}{DEAP} \\
\cmidrule{2-5}          & \multicolumn{1}{c}{Avg. Acc} & \multicolumn{1}{c|}{Avg. RA} & Avg. Acc & Avg. RA \\
    \midrule
    MIDA-EEG \cite{ZhengW2016} & $0.752\pm0.045$ & $0.843\pm0.020$ & $0.631\pm0.059$ & $0.743\pm0.037$ \\
    TCA-EEG \cite{LanZ2019}& $0.746\pm0.069$ & $0.852\pm0.038$ & $0.639\pm0.063$ & $0.746\pm0.041$ \\
    Deep-Transfer \cite{Tan2018} & $0.703\pm0.032$ & $0.829\pm0.017$ & $0.611\pm0.084$ & $0.732\pm0.056$ \\
    RA-MDRM \cite{Zanini2018} & $0.765\pm0.044$ & $0.857\pm0.028$ & $0.636\pm0.073$ & $0.751\pm0.045$ \\
    \midrule
    MUPS-EEG & $\boldsymbol{0.781\pm0.036}$ & $\boldsymbol{0.862\pm0.024}$ & $\boldsymbol{0.665\pm0.047}$ & $\boldsymbol{0.758\pm0.029}$ \\
    \bottomrule
    \end{tabular}%
  \label{t2}%
\end{table*}%

Table \ref{t2} reveals the model's capability to retain knowledge on previously learnt subjects after adaptation finishes. MUPS-EEG outperforms other models by a margin of 1.6\% on Avg. Acc and 0.5\% on Avg. RA for BCI-IV 2a dataset. For DEAP dataset, the model achieved a 2.6\% gain on Avg.Acc and 0.7\% gain on Avg.RA.  

\begin{figure*}[h!]
  \centering
  \begin{subfigure}[b]{0.45\textwidth}
  \includegraphics[width=\linewidth]{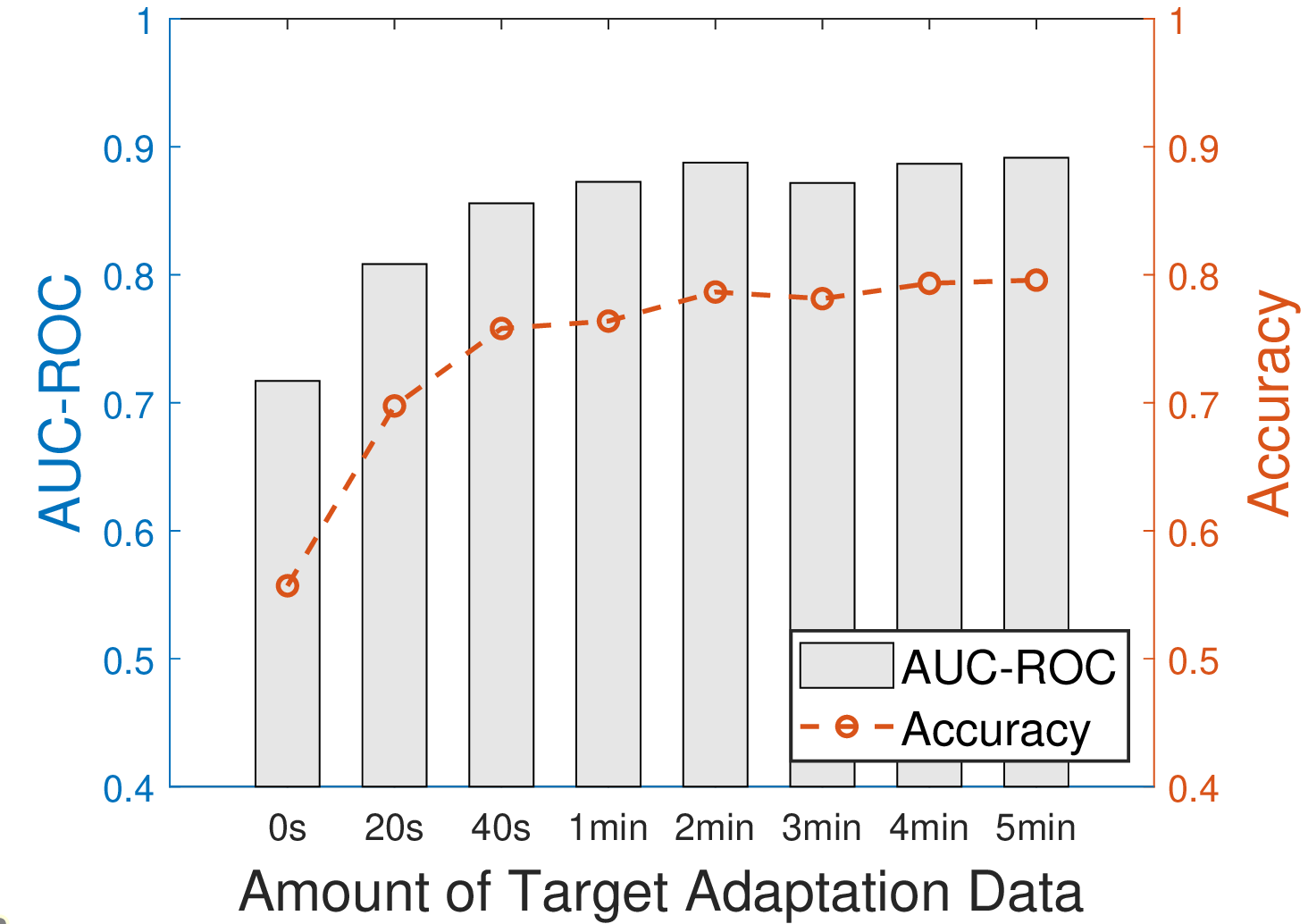}
  \caption{BCI IV-2a dataset}
  \label{f2_1}
  \end{subfigure}%
  \begin{subfigure}[b]{0.45\textwidth}
  \includegraphics[width=\linewidth]{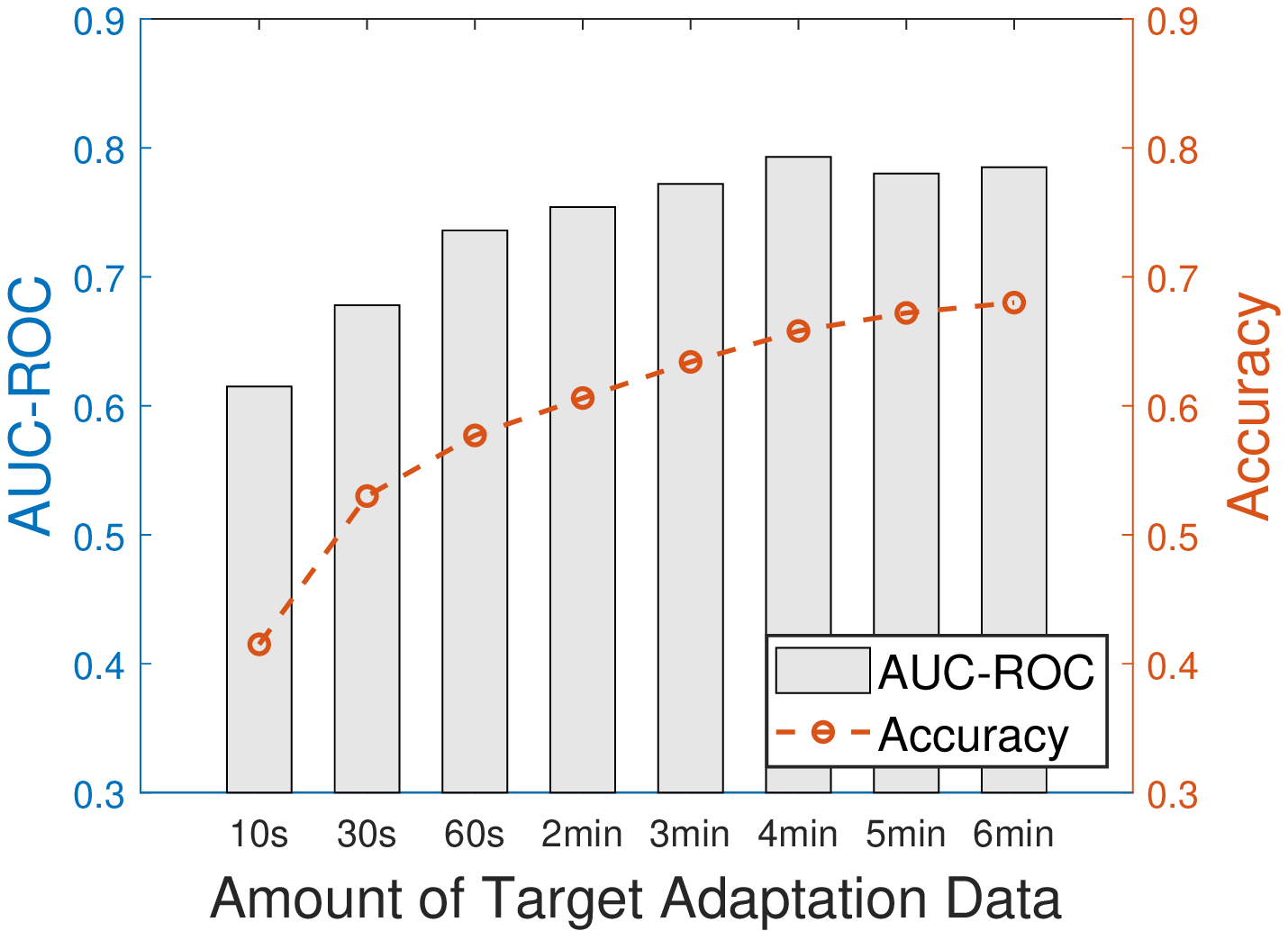}
  \caption{DEAP dataset}
  \label{f2_2}
  \end{subfigure}%
\caption{MUPS-EEG performance on (a) BCI IV-2a dataset and (b) DEAP dataset with different amount of target subject data. We observed both Accuracy and AUC-ROC score converge with 2 minutes of target subject data for BCI IV-2a dataset. 5 minutes of target data is needed for convergence on DEAP dataset.}
\label{f2}
\end{figure*}

We further explored the influence of different amount of target subject data on model performance, shown in fig. \ref{f2}. The performance is positively correlated with target data, and we observed both accuracy and AUC-ROC fully converges with 2 minutes of EEG recording from target subject on BCI IV-2a task, while 5 minutes of recording is needed for DEAP dataset. 

Comparing between model performance on the two datasets, DEAP posed to be more challenging than BCI IV-2a for the cross subject classification task, where only 3 out of 8 models reaches above 60\% accuracy in table \ref{t1}, given the theoretical chance for random guessing is 33.3\%. With current model performing below 70\% accuracy, which is generally deemed an acceptable threshold for application of BCI systems \cite{Fahimi2019}, further performance improvement is needed for DEAP dataset.

\section{Conclusion}
Pattern variability of EEG signal across different subjects is a major challenge for cross subject EEG classification. We propose an efficient transfer learning model built on meta update mechanism for the task. The two step meta update approach functioning on meta tasks enables the model to rapidly adapt onto a new subject and retain knowledge on known subjects at the same time. The model is efficient in terms of target data utilization with its tailored optimization process for target adaptation. We evaluate the model on two public datasets, where it outperforms current state of the arts by a large margin. 
%\nocite{*}
%\clearpage
\bibliographystyle{plain}
\balance
\bibliography{mups}

\end{document}